%% file: main.tex
\documentclass[10pt,twocolumn,letterpaper]{article}

\usepackage{wacv}
\input{preamble}

\definecolor{wacvblue}{rgb}{0.21,0.49,0.74}
\usepackage[pagebackref,breaklinks,colorlinks,allcolors=wacvblue]{hyperref}

\usepackage{multirow}
\usepackage{graphicx}
\usepackage{amsmath} 
\usepackage{amssymb}
\usepackage{float}
\usepackage{color, soul}
\usepackage{enumitem}
\usepackage{subcaption}
\usepackage{mathptmx} 
\usepackage[utf8]{inputenc} 
\usepackage[T1]{fontenc} 
\usepackage{booktabs}
\usepackage{subcaption}
\usepackage{dirtytalk}
\usepackage{adjustbox}
\usepackage{pgffor}
\usepackage{graphicx}
\usepackage{flafter}
\usepackage[section]{placeins}
\usepackage{makecell}
\usepackage{cuted} 

\title{Uncertainty-Aware Vision-Language Segmentation for Medical Imaging}

\author{
Aryan Das\thanks{Equal Contribution}\\
VIT Bhopal\\
{\tt\small aryan.das2021@vitbhopal.ac.in}
\and
Tanishq Rachamalla$^*$\\
SAHE, Andhra Pradesh\\
{\tt\small tanishqrachamalla12@gmail.com}
\and
Koushik Biswas\\
IIIT Delhi\\
{\tt\small koushikb@iiitd.ac.in}
\and
Swalpa Kumar Roy\\
Tezpur University Assam\\
{\tt\small swalpa@tezu.ernet.in}
\and
Vinay Kumar Verma\\
IIT Kanpur\\
{\tt\small vinayugc@gmail.com}
}

\begin{document}
\maketitle
\input{sec/0_abstract}    
\input{sec/1_intro}
\input{sec/2_related}
\input{sec/3_method}
\input{sec/4_experiments}
\input{sec/5_results}

\clearpage
{
    \small
    \bibliographystyle{IEEEtran}
    \bibliography{main}

}

\end{document}

%% file: preamble.tex
\definecolor{ForestGreen}{RGB}{34,139,34}
\definecolor{Blue}{RGB}{0,0,255}

%% file: sec/0_abstract.tex
\begin{abstract}
We introduce a novel uncertainty-aware multimodal segmentation framework that leverages both radiological images and associated clinical text for precise medical diagnosis. We propose a Modality Decoding Attention Block (MoDAB) with a lightweight State Space Mixer (SSMix) to enable efficient cross-modal fusion and long-range dependency modelling. To guide learning under ambiguity, we propose the Spectral-Entropic Uncertainty (SEU) Loss, which jointly captures spatial overlap, spectral consistency, and predictive uncertainty in a unified objective. In complex clinical circumstances with poor image quality, this formulation improves model reliability. Extensive experiments on various publicly available medical datasets, QATA-COVID19, MosMed++, and Kvasir-SEG, demonstrate that our method achieves superior segmentation performance while being significantly more computationally efficient than existing State-of-the-Art (SoTA) approaches. Our results highlight the importance of incorporating uncertainty modelling and structured modality alignment in vision-language medical segmentation tasks. Code: \href{https://github.com/arya-domain/UA-VLS}{https://github.com/arya-domain/UA-VLS}
\end{abstract}

%% file: sec/1_intro.tex
\vspace{-4mm}
\section{Introduction}
\label{sec:intro}

Medical image segmentation is a foundational task in computer-aided diagnosis, surgical planning, and clinical research~\cite{1, 3}. Deep learning has enabled automated image segmentation for assessing disease severity and guiding treatment. However, various unimodal methods depend heavily on extensive labelled data, which is often limited in clinical settings~\cite{2, 6}. To overcome this, recent studies have explored multimodal segmentation by integrating image data with textual reports. Leveraging natural language as auxiliary supervision offers rich contextual cues, enhancing segmentation performance, especially when visual quality is poor or annotations are sparse. 

Vision-language segmentation (VLS) aims to utilize natural language inputs, such as radiology reports or anatomical queries, to guide the segmentation process~\cite{12}. This multimodal paradigm offers several advantages: it mitigates the semantic disconnect between low-level visual cues and high-level clinical concepts, reduces the need for task-specific supervision, and enables more intuitive medical workflows~\cite{3, 15, 18}.

Despite progress in VLS, most existing methods neglect the role of uncertainty modelling during training, which is critical in clinical applications where predictions must be both accurate and reliable. Uncertainty-aware guidance can help models focus on ambiguous regions and reduce overconfident errors, especially when dealing with noisy data. However, uncertainty has largely been explored in unimodal medical segmentation, with minimal adoption in multimodal vision-language frameworks. Furthermore, effective alignment between visual features and language cues remains challenging, often limiting the benefits of cross-modal learning with limited parameters in a model. To address these issues, we incorporate an uncertainty-aware optimization and propose a state-space-based modality~\cite{mamba} integration strategy. This allows for efficient global dependency modelling while keeping the computational cost significantly lower than conventional transformer-based designs. Our contributions can be summarized as follows:
\begin{itemize}[left=1pt]
\item We propose Modality Decoding Attention Block (MoDAB) and State Space Mixer (SSMix) to enable structured multimodal fusion with long-range dependency modeling for medical vision-language tasks.
\item We also introduce Spectral-Entropic Uncertainty (SEU) Loss, a unified objective that integrates spatial, spectral, and uncertainty guidance into a single optimization.
\item Our computationally efficient model outperforms the recent State-of-The-Art (SoTA) methods on multiple benchmarks.
\end{itemize}

%% file: sec/2_related.tex
\section{Related Work}

\textbf{Unimodal Segmentation Models:} 
Early deep learning-based medical image segmentation models were largely built on fully convolutional networks, with U-Net~\cite{1} being the most influential. Enhanced variants like UNet++~\cite{2}, Attention U-Net~\cite{3}, and nnUNet~\cite{4} improved feature fusion via skip connections, dense pathways, and attention mechanisms. To capture global context, hybrid models emerged: TransUNet~\cite{5} combined CNNs with Vision Transformers (ViTs), and Swin-UNet~\cite{6} adopted hierarchical Swin Transformer blocks for multi-resolution processing. UCTransNet~\cite{7} further improved semantic understanding by integrating cross-fusion transformers and multi-head attention in skip connections. Sequence modelling approaches like U-Mamba~\cite{9} and Swin-UMamba~\cite{8} introduced Mamba-based modules into the U-Net, enabling long-range spatial dependency modelling through recurrent dynamics as an alternative to attention mechanisms.

\textbf{State Space Models:}  
State Space Models (SSMs) have emerged as promising alternatives to transformer-based architectures for long-sequence modeling due to their linear time complexity and memory efficiency. Gu et al.~\cite{gu2021efficiently} proposed S4, a structured state-space sequence model capable of capturing long-range dependencies while remaining computationally efficient. Subsequent advancements, including Hyena~\cite{poli2023hyena} and FlashAttention-2~\cite{dao2023flashattention2}, further demonstrated the effectiveness of structured memory mechanisms in sequence learning. Recently, Mamba~\cite{mamba} introduced selective state-space updates, enabling linear-time inference and training for long-range tasks with minimal compute overhead. While SSMs have shown strong results in language and vision domains, their application in multimodal and medical segmentation tasks remains limited.

\textbf{Vision-Language Segmentation Models:}
VLS has emerged as a transformative paradigm, enabling models to integrate clinical semantics with spatial reasoning for more interpretable and context-aware predictions. Foundational works like ConVIRT~\cite{10}, GLoRIA~\cite{14}, CLIP~\cite{12}, and BiomedCLIP~\cite{13} leveraged contrastive learning on paired medical images and textual reports, producing powerful joint embeddings that served as a backbone for a variety of downstream tasks. These models primarily focused on aligning vision and language representations at a global or hierarchical level, which laid the groundwork for segmentation models that could benefit from such multimodal understanding.
Building upon these pretrained foundations, transformer-based models such as ViLT~\cite{15}, LAVT~\cite{16}, and LViT-T~\cite{17} introduced mechanisms to directly inject textual information into the visual encoding pipeline using cross-modal attention, enabling dense prediction models to utilize linguistic prompts describing lesions, anatomical regions, or disease types. CMIRNet~\cite{27} advanced this paradigm by introducing sophisticated cross-modal interactive reasoning mechanisms specifically designed for referring image segmentation in medical contexts. Meanwhile, architectures like Ariadne~\cite{18} and SLViT~\cite{19} leveraged report-based supervision and multimodal attention to bridge the semantic gap in segmentation settings, using text as surrogate annotations to improve spatial localization.

More recent and specialized frameworks introduced tighter coupling between the two modalities; for example, TMCA~\cite{28} incorporated contrastive objectives at multiple levels of the network to improve feature alignment and semantic understanding across both modalities. Similarly, RecLMIS~\cite{22} employed a training paradigm where each modality helped reconstruct the other, encouraging shared latent understanding of anatomical and contextual features. MulModSeg~\cite{29} addressed the challenging problem of unpaired multi-modal medical image segmentation by introducing modality-conditioned text embedding and alternating training strategies. Likewise, DMMI~\cite{20} introduced dual-memory structures to separately capture and interact with visual and textual cues, reinforcing consistency and context awareness. Structured learning approaches such as RefSegformer~\cite{21} and TGANet~\cite{11} incorporated external references and graph-based language guidance, using textual anchors or graph attention mechanisms to disambiguate complex visual regions. Models like LGA~\cite{23} and TMC~\cite{36} pushed the limits of token-level and hierarchical language conditioning, embedding semantic meaning directly into the encoding and decoding stages of segmentation. Anatomical Structure-Guided Medical Vision-Language Pre-training represents a significant advancement in foundation model development by incorporating explicit anatomical structure awareness into the pre-training process. Finally, scalable models such as MAdapter~\cite{24} proposed language-guided adapters that could be inserted into vision transformers, enabling efficient multimodal learning without retraining the entire backbone. 

\textbf{Uncertainty in Medical Segmentation:}  
Uncertainty estimation has emerged as a critical component in medical image segmentation, particularly for enhancing model reliability in high-stakes clinical environments. Zeevi et al.~\cite{zeevi2025frequency} introduced Monte-Carlo Frequency Dropout (MC-FD), a technique that extends traditional MC-Dropout to the frequency domain. Their method demonstrated improved calibration and delineation of boundaries across diverse modalities, including MRI and CT. Similarly, Antico et al.~\cite{antico2022evaluating} performed a comprehensive evaluation of uncertainty quantification techniques across multiple algorithms and datasets, concluding that pixel-wise uncertainty estimation, especially using MC-Dropout, significantly improves the robustness and interpretability of segmentation models. Entropy is one of the most interpretable and computationally efficient measures of uncertainty. Sedai et al.~\cite{sedai2019uncertainty} utilized pixel-wise entropy maps to estimate aleatoric uncertainty in retinal vessel segmentation. Similarly, Roy et al.~\cite{roy2019bayesian} demonstrated the use of entropy from softmax outputs to capture both model and data uncertainty. These works highlight how entropy-based uncertainty can improve the interpretability of predictions and flag ambiguous regions, which is essential for downstream tasks.

%% file: sec/3_method.tex
\section{Methodology}

\begin{figure*}
    \vspace{-3mm}
    \centering
    \includegraphics[width=0.9\linewidth, height=0.45\linewidth]{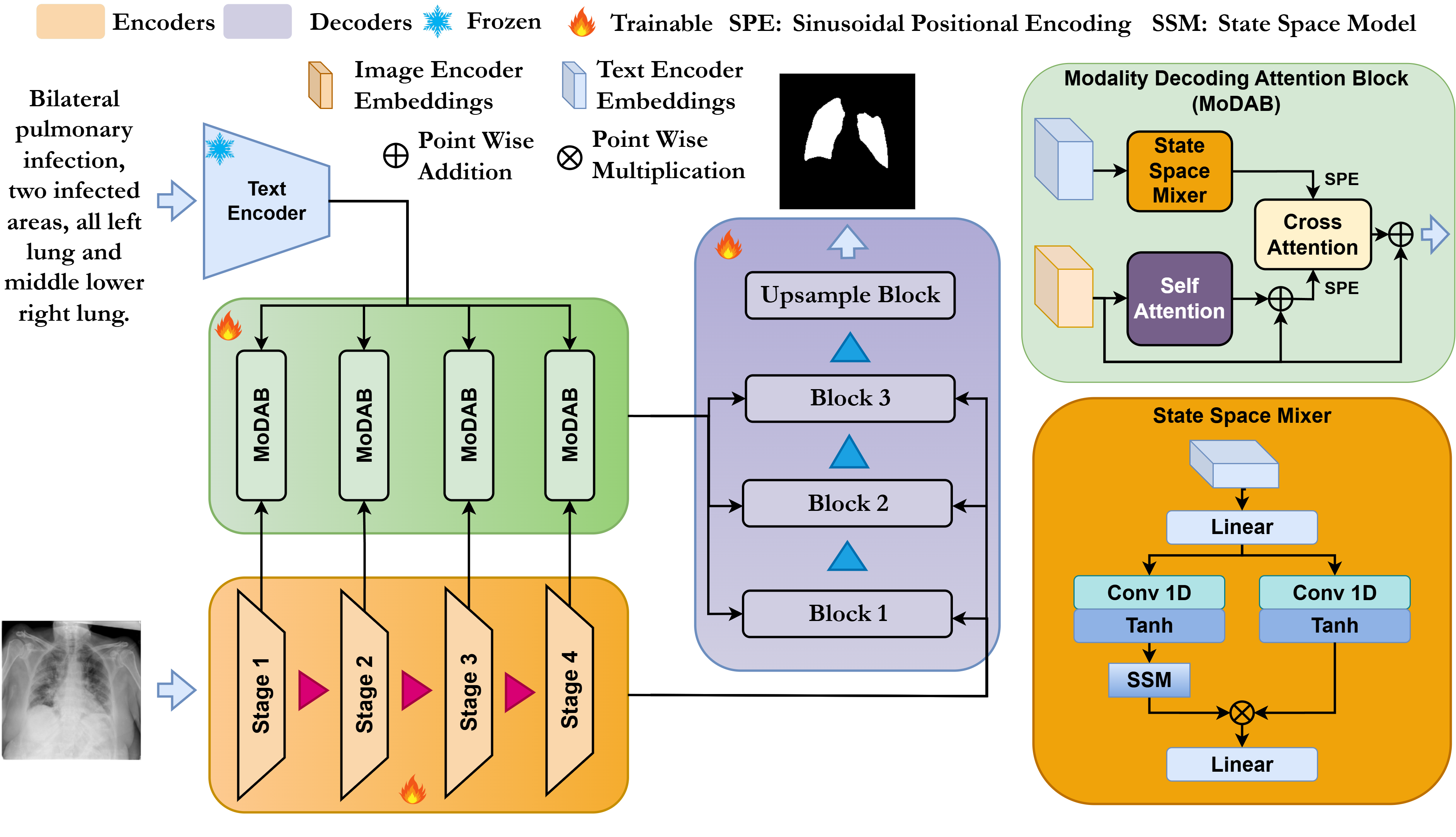}
    \vspace{-2mm}
    \caption{Overview of the proposed architecture. The model integrates visual and frozen text encoders and the Modality Decoding Attention Block (MoDAB), which incorporates Self-Attention and Cross-Attention along with a State Space Mixer (SSMix) for efficient multimodal fusion. The decoder reconstructs segmentation masks from the fused features through a multi-stage upsampling pathway.} 
    \label{fig:method}
    \vspace{-4mm}
\end{figure*}

\subsection{Modalities Encoding}

We utilize two pre-trained models to encode the input modalities: \textit{ConvNeXt-Tiny}~\cite{covnext} as the visual encoder and  \textit{BioViL CXR-BERT}~\cite{bert} as the text encoder. The visual encoder, denoted as $\mathcal{V}_\mathcal{E}$, extracts hierarchical features from four stages, capturing both fine-grained and abstract semantic information. Given a batch of chest X-ray images $\mathcal{I} \in \mathbb{R}^{B \times 3 \times H \times W}$ where $B$ is the batch size, $H$ is height and $W$ is the width, the visual encoder outputs a set of multi-scale feature maps:
\begin{equation}
   \mathcal{I'}_i  = \mathcal{V}_\mathcal{E}(\mathcal{I})
\end{equation}
where $\mathcal{I'}_i \in \mathbb{R}^{B \times C_i \times H_i \times W_i}$ denotes the feature map extracted at stage $i \in \{1, 2, 3, 4\}$. These features are spatially aligned and serve as inputs to subsequent modality decoding attention blocks.

For the textual input, we utilize a frozen text encoder, denoted as $\mathcal{T}_\mathcal{E}$, to extract contextualized token embeddings. Let the input token sequence be represented as $\mathcal{T} = [t_1, t_2, \dots, t_N]$, where $N$ denotes the sequence length. The encoder outputs a sequence of semantic embeddings:
\begin{equation}
    \mathcal{T'} = \mathcal{T}_\mathcal{E}(\mathcal{T})
\end{equation}
where $\mathcal{T'} \in \mathbb{R}^{B \times N \times D}$ denotes the output feature matrix, where each token is embedded in a $D$-dimensional space. These embeddings capture rich contextual semantics essential for cross-modal alignment.

\subsection{Modality Decoding Attention Block (MoDAB)}

The \textit{Modality Decoding Attention Block (MoDAB)} fuses spatial visual representations with contextual textual embeddings through a series of operations: Multi-Head Self-Attention ($SelfAttn$), Cross-Attention ($CrossAttn$) with Sinusoidal Positional Encodings (SPE), and a State Space Mixer (SSMix), which is a sequence mixer (detailed in Section~\ref{sec:ssmix}).

Let $\mathbf{X} \in \mathcal{I'}_i$ denote the visual input from the  $i^{th}$ stage of the visual encoder. The textual input $\mathcal{T'}$ is first projected to match the visual space via a learnable transformation, followed by a state-space-based mixer:
\begin{equation}
    \vspace{1mm}
    \mathcal{T}_{SSMix} = \textit{GELU}\left(\textit{SSMix}\left(\textit{LeakyReLU}\left(Linear(\mathcal{T'}) \right) \right) \right)
\end{equation}
 where $\mathcal{T}_{SSMix} \in \mathbb{R}^{B \times N \times Y_i}$, $B$ is the batch size and $Y_i = H_i \times W_i$ is the projected dimension.

\vspace{1mm}
\textbf{Self-Attention:}  
We apply \textit{Multi-Head Self Attention (MHSA)} to the visual sequence $\mathbf{X}$ to capture intra-modal dependencies among spatial tokens. First, the input is normalized ($LN$) and augmented with Sinusoidal Positional Encodings ($SPE$):
\begin{equation}
    \mathbf{X'} = \textit{SPE}\left( \textit{LN}(\mathbf{X}) \right)
\end{equation}
where $\mathbf{X'} \in \mathbb{R}^{B \times C_i \times Y_i}$ is linearly projected into $h$ attention heads, each with dimension $D_k$, using learned weight matrices:
\begin{equation}
    \mathbf{Q}_{SA_j} = \mathbf{X'} \mathbf{W}_{SA_j}^Q, \quad
    \mathbf{K}_{SA_j} = \mathbf{X'} \mathbf{W}_{SA_j}^K, \quad
    \mathbf{V}_{SA_j} = \mathbf{X'} \mathbf{W}_{SA_j}^V
\end{equation}
where $\mathbf{W}_{SA_j}^Q, \mathbf{W}_{SA_j}^K, \mathbf{W}_{SA_j}^V \in \mathbb{R}^{C_i \times D_k}$ for each head $j \in \{1, \dots, h\}$.

The attention for each head is computed using the scaled dot-product formulation:
\begin{equation}
    \textit{head}_j = \textit{SoftMax} \left( \frac{\mathbf{Q}_{SA_j} \mathbf{K}_{SA_j}^\top}{\sqrt{D_k}} \right) \mathbf{V}_{SA_j}
\end{equation}

All heads are concatenated and passed through a final projection:
\begin{equation}
    \textit{SelfAttn}(\mathbf{X'}) = \textit{Concat}(\textit{head}_1, \dots, \textit{head}_h) \mathbf{W}^O
\end{equation}
where $\mathbf{W}^O \in \mathbb{R}^{h \cdot D_k \times C}$ is a learned projection matrix. No masking is applied since all spatial tokens attend to each other. The MHSA output is normalized and added residually to form the self-attended feature:
\begin{equation}
    \mathbf{X_{SA}} = \mathbf{X'} + \textit{LN}\left( \textit{SelfAttn}(\mathbf{X'}) \right)
\end{equation}
where $\mathbf{X_{SA}} \in \mathbb{R}^{B \times C_i \times Y_i}$.

\vspace{1mm}
\textbf{Cross-Attention:}  
The \textit{Multi-Head Cross-Attention (MHCA)} extends MHSA by enabling cross-modal interaction: the query ($Q$) is derived from one modality, while the key ($K$) and value ($V$) come from another. Here, the self-attended visual features $\mathbf{X}_{SA}$ act as the query, and the state-space-enhanced textual embeddings $\mathcal{T}_{SSMix}$ provide the key and value. Similar to MHSA, both inputs are normalized and augmented with SPE to retain spatial and sequential structure.
\begin{align}
    \mathbf{Q}_{CA_j} &= \textit{SPE}\left( \textit{LN}(\mathbf{X}_{SA}) \right) \\
    \mathbf{K}_{CA_j} &= \textit{SPE}\left( \mathcal{T}_{SSMix} \right), \quad 
    \mathbf{V}_{CA_j} = \mathcal{T}_{SSMix}
\end{align}

The attention mechanism computes relevance between the visual queries and textual keys, producing cross-attended visual representations:
\begin{equation}
    \widehat{\mathbf{X}}_{\textbf{CA}} = \textit{CrossAttn}(\mathbf{Q}_{CA_j}, \mathbf{K}_{CA_j}, \mathbf{V}_{CA_j})
\end{equation}

To enable adaptive integration of textual context, the cross-attention output is normalized and added to the original visual features, scaled by a learnable scalar parameter $\alpha \in \mathbb{R}$:
\begin{equation}
    \mathbf{F} = \mathbf{X} + \alpha \cdot \textit{LN}\left(\widehat{\mathbf{X}}_{\textbf{CA}} \right)
\end{equation}
where $\alpha \in \mathbb{R}$ is randomly initialized and learned during training. $\mathbf{F} \in \mathbb{R}^{B \times C_i \times Y_i}$ captures both spatial visual dependencies and semantically aligned textual cues. This enriched feature map is subsequently propagated to the decoder for segmentation mask reconstruction.

\subsection{State Space Mixer (SSMix)}\label{sec:ssmix}

The \textit{State Space Mixer (SSMix)} is designed to enhance long-range dependency modelling in sequential data by combining learned temporal dynamics with convolutional operations and selective scanning mechanisms. Additionally, it incorporates a gating technique, serving as an efficient and lightweight module. Given the textual input $\mathcal{T'} \in \mathbb{R}^{B \times N \times D}$, the module outputs a transformed feature matrix $\mathcal{T}_{SSMix} \in \mathbb{R}^{B  \times N \times Y_i}$.
The input $\mathcal{T'}$ is first projected into an intermediate representation of size $2 \cdot d_{\text{inner}}$, where $d_{\text{inner}} = \gamma D$ and $\gamma$ is the expansion factor:
\begin{equation}
    \mathcal{T}_H = \textit{Linear}(\mathcal{T'}) \in \mathbb{R}^{B \times N \times 2d_{\text{inner}}}
\end{equation}

The projected features are then transposed to prepare for 1D convolution and split along the channel dimension into two parts:
\begin{equation}
    \mathbf{P}, \mathbf{Q} = \textit{Split}\left( \textit{Transpose}(\mathcal{T}_H, [0, 2, 1]) \right), \quad 
\end{equation}
where $\mathbf{P}, \mathbf{Q} \in \mathbb{R}^{B \times d_{\text{inner}} \times N}$.

After splitting the features, depthwise 1D convolutions are applied to each component to extract localized temporal features:
\begin{equation}
    \tilde{\mathbf{P}} = \textit{tanh}(\textit{Conv1D}_x(\mathbf{P})), \quad 
    \tilde{\mathbf{Q}} = \textit{tanh}(\textit{Conv1D}_z(\mathbf{Q}))
\end{equation}

The output $\tilde{\mathbf{P}}$ is passed through a linear projection to produce dynamic time-step parameters $\boldsymbol{\Delta}$ and state parameters $\mathbf{B}, \mathbf{C}$:
\begin{equation}
    [\boldsymbol{\Delta}, \mathbf{B}, \mathbf{C}] = \textit{Split}\left( \textit{Linear}(\tilde{\mathbf{P}}) \right)
\end{equation}

The stepping weights $\boldsymbol{\Delta} \in \mathbb{R}^{B \times d_{inner} \times L}$ are further refined through a softplus reparameterization of their log-transformed initializations to ensure stability: 
\begin{equation}
    \Delta = \textit{Softplus}(\boldsymbol{\Delta} + \text{bias}_\Delta)
\end{equation}

A state-space update is then performed using the \textit{selective State Space Model (SSM)}, which models latent dynamics across time steps with exponentially decaying memory kernels. 
The state output $\mathbf{SCAN}$ is computed as:
\begin{equation}
    \mathbf{SCAN} = \textit{SSM}(\tilde{\mathbf{X}}, \Delta, \mathbf{A}, \mathbf{B}, \mathbf{C}, \mathbf{E})
\end{equation}
where $\mathbf{E}$ is a learned gating vector applied to modulate the scan dynamics and $\mathbf{A}$ is a diagonal state transition matrix.

Finally, the output $\mathbf{SCAN}$ is concatenated with the convolutional branch $\tilde{\mathbf{Q}}$, and the result is projected back to the output embedding dimension using a final linear transformation:
\begin{equation}
    \mathcal{T}_{SSMix} = \textit{Linear} \left( \textit{Concat}(\mathbf{SCAN}, \tilde{\mathbf{Q}})^\top \right)
\end{equation}

The resulting $\mathcal{T}_{SSMix}$ captures both global and local dependencies, facilitating effective multimodal fusion in downstream decoding.

\subsection{Decoder}

The decoder reconstructs the spatial segmentation layout by first reshaping the fused multimodal feature $\mathbf{F} \in \mathbb{R}^{B \times C_i \times Y_i}$ into a spatial feature map $\mathbf{F'} \in \mathbb{R}^{B \times C_i \times H_i \times W_i}$. It follows a four-stage decoding pipeline that progressively restores the spatial resolution.
For each stage $m \in \{1, 2, 3\}$, an \textit{Upsampling Block} doubles the spatial resolution using a transposed convolution operation:
\begin{equation}
    \mathbf{F}^{(m)}_{\text{up}} = \text{TransConv}(\mathbf{F}^{(m-1)}), \quad \mathbf{F}^{(0)} := \mathbf{F'}
\end{equation}
where $\text{TransConv}(\cdot)$ denotes a $2 \times 2$ transposed convolution with stride 2. The upsampled feature map $\mathbf{F}^{(m)}_{\text{up}} \in \mathbb{R}^{B \times C_m \times H_m \times W_m}$ captures progressively finer spatial structure, with $C_m$ representing the output channels at stage $m$.

The upsampled feature $\mathbf{F}^{(m)}_{\text{up}}$ is concatenated with the corresponding encoder feature $\mathcal{I'}_{4-m}$ at the same resolution level. The resulting tensor is processed by a \textit{Convolutional Refinement Block (CRB)} $\mathit{CRB}_m$, comprising two convolutional layers, LeakyReLU activations, and batch normalization:
\begin{equation}
    \mathbf{F}_{\text{CRB}}^{m} = \textit{CRB}_m\left( \textit{Concat}(\mathbf{F}^{(m)}_{\text{up}}, \mathcal{I'}_{4-m}) \right)
\end{equation}

The final stage applies a \textit{Subpixel Upsampling Network (SUN)}, consisting of a convolutional layer followed by pixel shuffling. The convolution increases the feature dimensionality:
\begin{equation}
    \mathbf{F}_{\text{pre}} = \textit{Conv2D}(\mathbf{F}_{\text{CRB}}^{3})
\end{equation}
Pixel shuffling $\Pi(\cdot)$ rearranges spatial elements to produce a high-resolution output by a factor of $r$ in each spatial dimension:
\begin{equation}
    \mathbf{F}_{\text{SU}} = \Pi(\mathbf{F}_{\text{pre}})
\end{equation}
yielding $\mathbf{F}_{\text{SU}} \in \mathbb{R}^{B \times C \times rH \times rW}$.

To improve local consistency and mitigate boundary artifacts, an average pooling operation with $o \times o$ kernel and appropriate padding is applied:
\begin{equation}
    \mathbf{F}_{\text{avg}} = \textit{AvgPool2D}(\textit{Pad}(\mathbf{F}_{\text{SU}}))
\end{equation}

Finally, a $1 \times 1$ convolutional output layer maps the refined features into the desired number of prediction channels:
\begin{equation}
    \hat{\mathbf{Y}} = \textit{Conv}_{1\times1}(\mathbf{F}_{\text{avg}}) \in \mathbb{R}^{B \times C_o \times H \times W}
\end{equation}
where $C_o$ is the number of output channels. This multi-stage decoding process enables coarse-to-fine segmentation reconstruction, preserving both semantic and spatial detail through visual-textual alignment.

\subsection{Objective Function}

To guide the model toward anatomically precise, structurally consistent, and uncertainty-aware predictions, we introduce the \textit{Spectral-Entropic Uncertainty (SEU) Loss}, a unified objective designed for medical vision-language segmentation. Rather than treating separate objectives independently, SEU Loss holistically integrates spatial, spectral, and probabilistic priors into a single formulation.

Let $\hat{\mathbf{Y}} \in \mathbb{R}^{B \times C \times H \times W}$ denote the predicted segmentation map and $\hat{\mathbf{G}} \in \mathbb{R}^{B \times C \times H \times W}$ the one-hot encoded ground truth. The SEU loss is expressed as:
\begin{equation}
\begin{split}
\mathcal{L}{\text{SEU}} =\; & \mathcal{L}{\text{Dice}}(\hat{\mathbf{Y}}, \hat{\mathbf{G}})
+ \lambda_{\text{F}} \cdot \mathcal{R}{\text{Spectral}}(\hat{\mathbf{Y}}, \hat{\mathbf{G}}) \\
& + \lambda_{\text{E}} \cdot \mathcal{R}{\text{Entropy}}(\hat{\mathbf{Y}})
\end{split}
\end{equation}
where $\lambda_{\text{F}}$ and $\lambda_{\text{E}}$ are modulation weights for spectral alignment and uncertainty regularization, respectively. Each component contributes to a different representational aspect, but collectively they form a single landscape.

\vspace{1mm}
\textbf{Spatial Alignment:}
The core supervision comes from a differentiable Dice loss, capturing the pixel-level overlap between $\hat{\mathbf{Y}}$ and $\hat{\mathbf{G}}$:
\begin{equation}
\mathcal{L}_{\text{Dice}} = 1 - \frac{2 \cdot \sum (\hat{\mathbf{Y}} \cdot \hat{\mathbf{G}}) + \epsilon}{\sum \hat{\mathbf{Y}} + \sum \hat{\mathbf{G}} + \epsilon}
\end{equation}
where the summation is over all spatial and channel dimensions, and $\epsilon$ is a small constant for numerical stability.

\vspace{1mm}
\textbf{Spectral Consistency:}
To enforce global structural fidelity, we align the magnitude of Fourier spectra between the predicted and target masks:
\begin{equation}
\mathcal{R}_{\text{Spectral}} = \left| \left| \mathcal{F}(\hat{\mathbf{Y}}) \right| - \left| \mathcal{F}(\hat{\mathbf{G}}) \right| \right|_2^2
\end{equation}
where $\mathcal{F}(\cdot)$ denotes the 2D Fourier Transform and $|\cdot|$ is the magnitude operation. This encourages preservation of global anatomical topology, especially beneficial for diffuse or subtle lesions.

\vspace{1mm}
\textbf{Uncertainty Guidance:}  
To penalize ambiguous predictions and promote confident outputs, we incorporate an entropy-based regularization term defined as:
\begin{equation}
    \mathcal{R}_{\text{entropy}} = -\frac{1}{BHW} \sum_{b,c,h,w} \hat{\mathbf{Y}}_{b,c,h,w} \log(\hat{\mathbf{Y}}_{b,c,h,w} + \delta)
\end{equation}
where the indices \( b \), \( c \), \( h \), and \( w \) range over \( b \in \{1, \dots, B\} \), \( c \in \{1, \dots, C\} \), \( h \in \{1, \dots, H\} \), and \( w \in \{1, \dots, W\} \), corresponding to the batch size, number of classes, and spatial dimensions (height and width), respectively. The term \( \delta \) is a small constant added for numerical stability to prevent undefined values. This entropy-based regularization term acts as a soft constraint, encouraging the model to reduce uncertainty by promoting low-entropy, confident predictions.

%% file: sec/4_experiments.tex
\section{Experiments}

\begin{figure*}[t]
    \vspace{-3mm}
    \centering
    \includegraphics[width=0.85\linewidth, height=0.28\textheight]{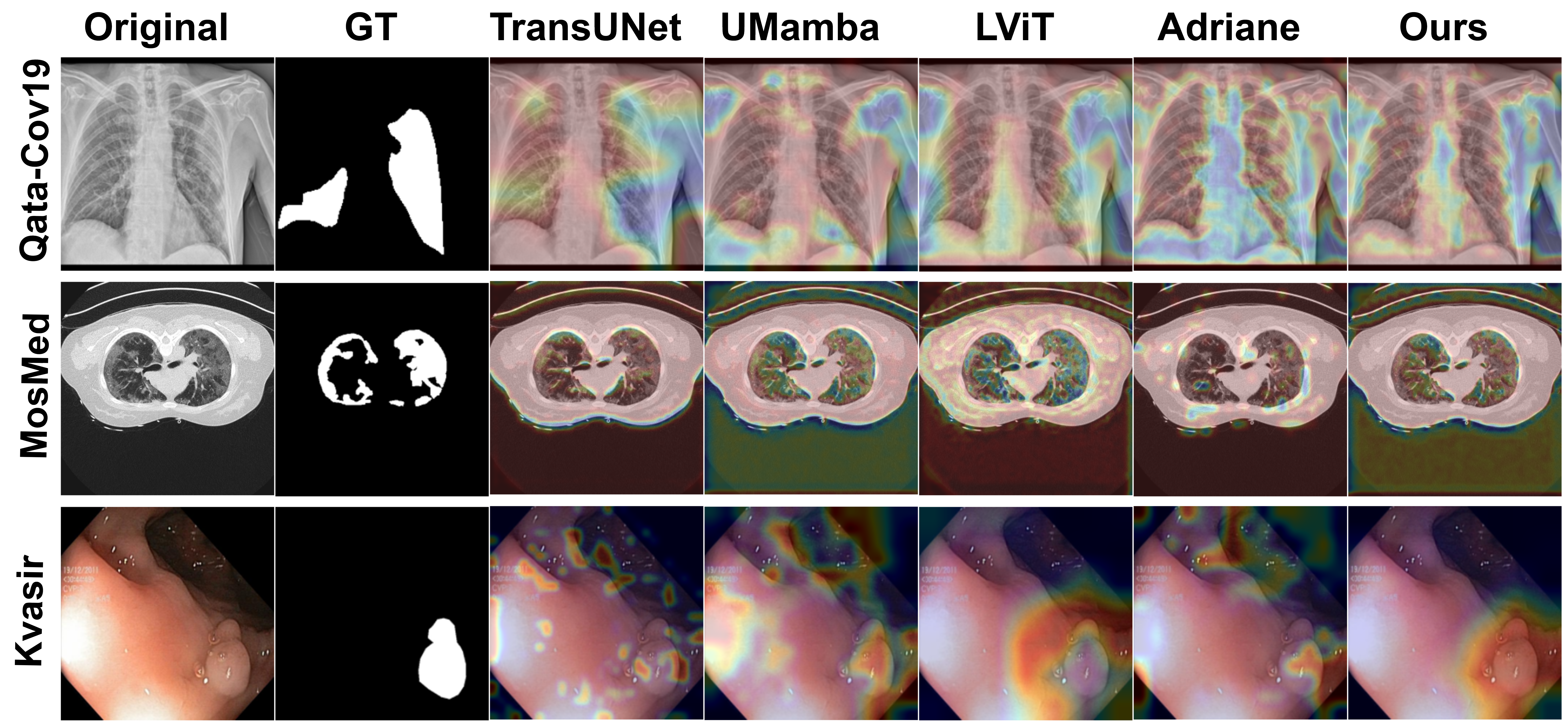}
    \vspace{-1mm}
    \caption{Comparison of Grad-CAM-Based Attention Visualizations Between the Proposed Model and Baseline methods}
    \label{fig:grad_cam}
    \vspace{-3mm}
\end{figure*}

\begin{figure*}[t]
    \centering
    \includegraphics[width=0.90\linewidth, height=0.28\textheight]{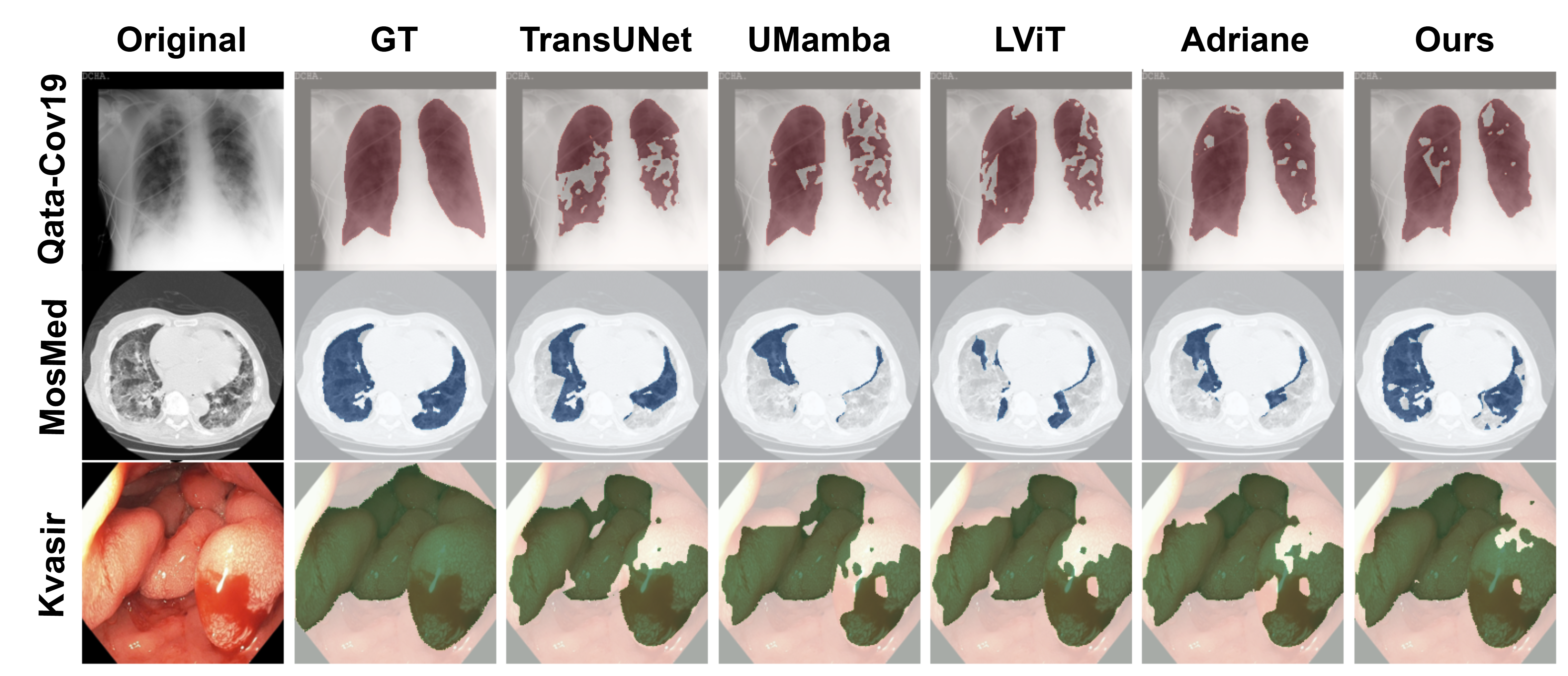}
    \vspace{-1mm}
    \caption{Qualitative Comparison of Predicted Segmentation Maps with Baseline Models}
    \label{fig:visualization}
    \vspace{-4mm}
\end{figure*}

\input{sec/tables}

\subsection{Experimental Setup}
All experiments were carried out on a high-performance computing server equipped with an \textit{Intel Xeon Silver 4214R} CPU running at 2.40 GHz, 128 GB of RAM, and \textit{NVIDIA A30} GPUs. The system environment was configured with \textit{CUDA version 12.2} to ensure GPU acceleration for training and evaluation. 

\subsection{Datasets}

We employ three publicly available datasets: \textbf{QaTa-COV19}, \textbf{MosMed++}, and \textbf{Kvasir-SEG} to evaluate the proposed model. These datasets, initially developed for unimodal segmentation purposes, have been augmented with concise natural language descriptions by recent works such as LViT~\cite{17} and MedVLSM~\cite{medvlsm}, enabling vision-language segmentation. 

\textbf{QaTa-COV19:}  
The QaTa-COV19 dataset~\cite{48degerli2022osegnet}, compiled by researchers from Qatar University and Tampere University, comprises 9,258 chest X-ray images of COVID-19 cases. It is one of the first datasets to include manually annotated COVID-19 lesion regions. To support multimodal training, this dataset was extended with textual descriptions by LViT~\cite{17}. 

\textbf{MosMed++:}  
MosMed++~\cite{morozov2020mosmeddata, hofmanninger2020automatic} is a chest CT dataset containing 2,729 axial slices from patients diagnosed with COVID-19. Each slice is annotated with severity scores and enriched with textual descriptions curated by LViT~\cite{17}. 

\textbf{Kvasir-SEG:}  
The Kvasir-SEG dataset~\cite{kvasir_dataset} includes 1,000 high-resolution gastrointestinal endoscopy images with pixel-wise polyp annotations. The image resolutions vary from 332×487 to 1920×1072 pixels. MedVLSM~\cite{medvlsm} introduced caption annotations describing polyp characteristics. For consistency with other datasets, we selected a single caption per image, prioritizing those that included location-based descriptors. 

\subsection{Training Details}

Experiments were conducted using a consistent training and validation setup across all three datasets. Input images were uniformly resized to a resolution of $224 \times 224$ pixels to ensure compatibility with the model architecture and to standardize training across datasets, thereby maintaining consistency in spatial features. A batch size of 32 was used during both training and validation phases. The training process was executed for a maximum of 200 epochs, with early stopping implemented based on a patience of 20 epochs to avoid overfitting and ensure generalizability. A minimum training duration was enforced, requiring at least 20 epochs to promote model stability during the initial learning phase. {The values of $\lambda_{\text{F}}$ and $\lambda_{\text{E}}$ are $0.3$ and $0.1$, respectively.}

We also employed the AdamW optimizer~\cite{loshchilov2019decoupled}, which decouples weight decay from the gradient update process, making it particularly effective for transformer-based architectures. The initial learning rate was set to $5 \times 10^{-4}$ for the MosMed++ dataset and $3 \times 10^{-4}$ for the QaTa-COV19 and Kvasir-SEG datasets, based on preliminary tuning and empirical observations for optimal convergence. Additionally, a cosine annealing learning rate scheduler~\cite{loshchilov2016sgdr} was utilized to progressively reduce the learning rate over time, with a maximum cycle length of 200 epochs and a minimum learning rate threshold of $1 \times 10^{-6}$.

%% file: sec/tables.tex
\begin{table*}[!t]
\centering
\caption{Comparison of Monomodal and Multimodal State-of-The-Art (SoTA) methods on medical image segmentation across three datasets: QaTa-COV19, MosMed++, and Kvasir-SEG. Metrics include Dice score (\%), mean Intersection over Union (mIoU, \%), number of Trainable Parameters (Millions), and Floating-Point Operations (FLOPs) per second (Billions). \textbf{Black}: best, \textbf{\textcolor{ForestGreen}{Green}}: second best, \textbf{\textcolor{Blue}{Blue}}: third best values.}
\vspace{-2mm}
\label{tab:results}
\resizebox{\textwidth}{!}{%
\begin{tabular}{@{}|c|c|c|c|cc|cc|cc|@{}}
\toprule
\multirow{2}{*}{\textbf{Modality}} & \multirow{2}{*}{\textbf{Method}} & \multirow{2}{*}{\textbf{Trainable Params (M)}} & \multirow{2}{*}{\textbf{Flops (G)}} & \multicolumn{2}{c|}{\textbf{QATA-COV19}} & \multicolumn{2}{c|}{\textbf{MosMedData++}} & \multicolumn{2}{c|}{\textbf{Kvasir-Seg}} \\ \cmidrule(l){5-10}
& & & & \multicolumn{1}{c|}{\textbf{Dice (\%)}} & \textbf{mIoU (\%)} & \multicolumn{1}{c|}{\textbf{Dice (\%)}} & \textbf{mIoU (\%)} & \multicolumn{1}{c|}{\textbf{Dice (\%)}} & \textbf{mIoU (\%)} \\ \midrule
\multirow{9}{*}{\rotatebox{90}{\textbf{MonoModels}}} & U-Net~\cite{1} & 14.8 & 50.3 & \multicolumn{1}{c|}{78.91} & 69.32 & \multicolumn{1}{c|}{64.30} & 50.50 & \multicolumn{1}{c|}{82.33} & 74.26 \\ \cmidrule(l){2-10}
& UNet++~\cite{2} & 74.5 & 94.6 & \multicolumn{1}{c|}{79.47} & 70.05 & \multicolumn{1}{c|}{71.63} & 58.14 & \multicolumn{1}{c|}{82.79} & 73.94 \\ \cmidrule(l){2-10}
& AttUNet~\cite{3} & 34.9 & 101.9 & \multicolumn{1}{c|}{79.11} & 69.83 & \multicolumn{1}{c|}{66.07} & 52.68 & \multicolumn{1}{c|}{82.94} & 74.17 \\ \cmidrule(l){2-10}
& nnUNet~\cite{4} & 19.1 & 412.7 & \multicolumn{1}{c|}{80.30} & 70.62 & \multicolumn{1}{c|}{72.32} & 60.14 & \multicolumn{1}{c|}{84.22} & 75.41 \\ \cmidrule(l){2-10}
& TransUNet~\cite{5} & 105 & 56.7 & \multicolumn{1}{c|}{78.44} & 68.84 & \multicolumn{1}{c|}{71.13} & 58.28 & \multicolumn{1}{c|}{90.53} & 85.94 \\ \cmidrule(l){2-10}
& Swin-Unet~\cite{6} & 82.3 & 67.3 & \multicolumn{1}{c|}{77.85} & 68.07 & \multicolumn{1}{c|}{63.19} & 49.93 & \multicolumn{1}{c|}{89.09} & 85.84 \\ \cmidrule(l){2-10}
& UCTransNet~\cite{7} & 65.6 & 63.2 & \multicolumn{1}{c|}{79.00} & 69.34 & \multicolumn{1}{c|}{65.71} & 52.55 & \multicolumn{1}{c|}{91.04} & 87.31 \\ \cmidrule(l){2-10}
& Swin-UMamba~\cite{8} & 60 & 68 & \multicolumn{1}{c|}{80.02} & 70.11 & \multicolumn{1}{c|}{65.31} & 51.28 & \multicolumn{1}{c|}{79.43} & 68.63 \\ \cmidrule(l){2-10}
& U-Mamba~\cite{9} & 18.51 & 375.78 & \multicolumn{1}{c|}{80.51} & 70.89 & \multicolumn{1}{c|}{65.88} & 52.17 & \multicolumn{1}{c|}{89.81} & 85.14 \\ \midrule
\multirow{15}{*}{\rotatebox{90}{\textbf{Multimodal}}} & ConVIRT~\cite{10} & 35.2 & 44.6 & \multicolumn{1}{c|}{79.45} & 70.29 & \multicolumn{1}{c|}{71.92} & 59.52 & \multicolumn{1}{c|}{89.24} & 83.01 \\ \cmidrule(l){2-10}
& TGANet~\cite{11} & \textbf{\textcolor{ForestGreen}{19.8}} & 41.9 & \multicolumn{1}{c|}{79.66} & 70.61 & \multicolumn{1}{c|}{71.63} & 59.00 & \multicolumn{1}{c|}{89.76} & 83.20 \\ \cmidrule(l){2-10}
& CLIP~\cite{12} & 87 & 105.3 & \multicolumn{1}{c|}{79.57} & 70.54 & \multicolumn{1}{c|}{71.75} & 59.43 & \multicolumn{1}{c|}{90.04} & 86.29 \\ \cmidrule(l){2-10}
& BiomedClip~\cite{13} & 87 & 105.3 & \multicolumn{1}{c|}{87.75} & 78.10 & \multicolumn{1}{c|}{66.33} & 50.27 & \multicolumn{1}{c|}{85.34} & 77.60 \\ \cmidrule(l){2-10}
& GLoRIA~\cite{14} & 45.6 & 60.8 & \multicolumn{1}{c|}{79.83} & 70.54 & \multicolumn{1}{c|}{72.36} & 60.12 & \multicolumn{1}{c|}{86.10} & 77.93 \\ \cmidrule(l){2-10}
& ViLT~\cite{15} & 87.4 & 55.9 & \multicolumn{1}{c|}{79.45} & 70.02 & \multicolumn{1}{c|}{72.10} & 59.85 & \multicolumn{1}{c|}{86.42} & 76.91 \\ \cmidrule(l){2-10}
& LAVT~\cite{16} & 118.6 & 83.8 & \multicolumn{1}{c|}{79.15} & 69.73 & \multicolumn{1}{c|}{73.10} & 60.14 & \multicolumn{1}{c|}{87.16} & 74.72 \\ \cmidrule(l){2-10}
& LViT~\cite{17} & 29.7 & 54.1 & \multicolumn{1}{c|}{83.40} & 74.89 & \multicolumn{1}{c|}{74.32} & 61.03 & \multicolumn{1}{c|}{87.59} & 75.16 \\ \cmidrule(l){2-10}
& Ariadne~\cite{18} & 43.95 & \textbf{\textcolor{ForestGreen}{22.36}} & \multicolumn{1}{c|}{\textbf{\textcolor{Blue}{88.06}}} & \textbf{\textcolor{Blue}{79.00}} & \multicolumn{1}{c|}{\textbf{\textcolor{Blue}{78.29}}} & 64.44 & \multicolumn{1}{c|}{90.45} & 82.34 \\ \cmidrule(l){2-10}
& SLViT~\cite{19} & 131.5 & 51.1 & \multicolumn{1}{c|}{79.10} & 68.71 & \multicolumn{1}{c|}{72.36} & 60.55 & \multicolumn{1}{c|}{89.88} & 83.03 \\ \cmidrule(l){2-10}
& DMMI~\cite{20} & 114.6 & 63.3 & \multicolumn{1}{c|}{83.85} & 75.42 & \multicolumn{1}{c|}{74.78} & 61.59 & \multicolumn{1}{c|}{\textbf{\textcolor{Blue}{90.96}}} & 83.41 \\ \cmidrule(l){2-10}
& RefSegformer~\cite{21} & 195 & 103.6 & \multicolumn{1}{c|}{83.93} & 75.34 & \multicolumn{1}{c|}{74.81} & 61.46 & \multicolumn{1}{c|}{90.57} & \textbf{\textcolor{Blue}{83.69}} \\ \cmidrule(l){2-10}
& RecLMIS~\cite{22} & \textbf{\textcolor{Blue}{23.7}} & \textbf{\textcolor{Blue}{24.1}} & \multicolumn{1}{c|}{84.93} & 76.86 & \multicolumn{1}{c|}{77.26} & \textbf{\textcolor{Blue}{64.95}} & \multicolumn{1}{c|}{86.58} & 77.08 \\ \cmidrule(l){2-10}
& LGA~\cite{23} & \textbf{8.24} & 382.17 & \multicolumn{1}{c|}{84.40} & 76.05 & \multicolumn{1}{c|}{62.30} & \textbf{75.43} & \multicolumn{1}{c|}{89.82} & 83.25 \\ \cmidrule(l){2-10}
& MAdapter~\cite{24} & - & - & \multicolumn{1}{c|}{\textbf{\textcolor{ForestGreen}{90.07}}} & \textbf{\textcolor{ForestGreen}{81.88}} & \multicolumn{1}{c|}{\textbf{\textcolor{ForestGreen}{78.40}}} & 62.77 & \multicolumn{1}{c|}{\textbf{\textcolor{ForestGreen}{91.37}}} & \textbf{\textcolor{ForestGreen}{84.36}} \\ \cmidrule(l){2-10}
& \textbf{Our Model} & 39.9 & \textbf{17.87} & \multicolumn{1}{c|}{\textbf{92.24}} & \textbf{84.9} & \multicolumn{1}{c|}{\textbf{79.67}} & \textbf{\textcolor{ForestGreen}{66.38}} & \multicolumn{1}{c|}{\textbf{93.83}} & \textbf{87.62} \\ \bottomrule
\end{tabular}%
}
\vspace{-4mm}
\end{table*}

%% file: sec/5_results.tex
\section{Results and Discussion}

\subsection{Qualitative and Quantitative Analysis}
As shown in Figure~\ref{fig:grad_cam}, our model exhibits more focused and semantically aligned attention compared to SoTAs. In Figure~\ref{fig:visualization}, we present qualitative segmentation results on three datasets, highlighting only the main predicted regions for clarity. Our model demonstrates superior precision in localizing and delineating the target areas compared to SoTAs.

Our proposed model demonstrates superior performance across all datasets, achieving state-of-the-art results in both Dice coefficient and mean Intersection over Union (mIoU) metrics (Table~\ref{tab:results}). On the QATA-COV19 dataset, the top-performing baselines include MAdapter (90.07\% Dice, 81.88\% mIoU), BiomedClip (87.75\% Dice, 78.10\% mIoU), and RecLMIS (84.93\% Dice, 76.86\% mIoU) among multimodal approaches, while U-Mamba (80.51\% Dice, 70.89\% mIoU) and nnUNet (80.30\% Dice, 70.62\% mIoU) represent the best monomodal methods. Our model achieved exceptional performance with a Dice score of 92.24\% and mIoU of 84.9\%, demonstrating +2.17\% improvement over MAdapter, +4.49\% over BiomedClip, and \textbf{+11.73\%} over the best monomodal approach, U-Mamba.

On the MosMedData++ dataset, the top-performing baselines include MAdapter (78.40\% Dice, 62.77\% mIoU), Ariadne (78.29\% Dice, 64.44\% mIoU), and RecLMIS (77.26\% Dice, 64.95\% mIoU) among multimodal approaches, while nnUNet (72.32\% Dice, 60.14\% mIoU) and UNet++ (71.63\% Dice, 58.14\% mIoU) represent the best monomodal methods. Our model achieved a Dice score of 79.67\% and mIoU of 66.38\%, establishing new state-of-the-art results with +1.27\% improvement over MAdapter, +1.38\% over Ariadne, and +7.35\% over the best monomodal method, nnUNet.

On the Kvasir-Seg dataset for polyp segmentation, the top-performing baselines include MAdapter (91.37\% Dice, 84.36\% mIoU), UCTransNet (91.04\% Dice, 87.31\% mIoU), and DMMI (90.96\% Dice, 83.41\% mIoU) among multimodal approaches, while TransUNet (90.53\% Dice, 85.94\% mIoU) and U-Mamba (89.81\% Dice, 85.14\% mIoU) represent the best monomodal methods. Our model achieved outstanding performance with a Dice score of 93.83\% and mIoU of 87.62\%, demonstrating +2.46\% improvement over MAdapter, +2.79\% over UCTransNet, and +3.3\% over the best monomodal approach, TransUNet.

\subsection{Computational Efficiency Analysis}
Our proposed model demonstrates remarkable computational efficiency while maintaining superior performance, as shown in Table~\ref{tab:results}. With only 39.9M trainable parameters, our model is significantly more compact than many SoTAs, such as RefSegformer (195M), SLViT (131.5M), and LAVT (118.6M). Our model requires only 17.87G FLOPs, making it the most efficient in the SoTAs. Despite being more efficient than most baselines, our model consistently achieves the highest performance across all datasets, demonstrating an excellent performance-efficiency trade-off.

\vspace{-1mm}
\section{Ablation Studies}
\begin{table}[t]
    \small
    \centering
    \caption{Ablation results on the \textbf{Kvasir-SEG} dataset. Each condition evaluates the model after removing or replacing a key component. Metrics reported are Dice score (\%) and mIoU (\%).}
    \vspace{-2mm}
    \label{tab:ablation}
    \begin{tabular}{p{4.5cm}|c| c}
        \toprule
        \textbf{Method} & \textbf{Dice (\%)} & \textbf{mIoU (\%)} \\
        \midrule
        \multicolumn{3}{l}{\textit{Loss Function}} \\
        \quad Dice loss & 93.44 & 85.76 \\
        \quad BCE loss  & 92.03 & 85.26 \\
        \midrule
        \multicolumn{3}{l}{\textit{Textual Guidance}} \\
        \quad Inference w/o Text Prompts    & 87.28 & 81.32 \\
        \quad Training w/o MoDAB  & 85.15 & 73.86 \\
        \midrule
        \multicolumn{3}{l}{\textit{Architectural Replacements}} \\
        \quad  SSMix with Linear layer         & 91.72 & 82.43 \\
        \quad  Cross-Attention with Addition   & 92.11 & 82.59 \\
        \midrule
        \textbf{Complete Model (ours)}              & \textbf{93.86} & \textbf{87.62} \\
        \bottomrule
    \end{tabular}
    \vspace{-4mm}
\end{table}

To assess the contributions of key components in our framework, we conduct a detailed ablation study using the Kvasir-SEG dataset. The experiments are categorized into three aspects: Loss formulation, Textual Guidance, and Architectural Components. Results in Table~\ref{tab:ablation} highlight the performance drop associated with the removal or replacement of specific modules, thereby validating their necessity.

\textbf{Loss Function Analysis:}  
We evaluate the impact of our proposed Spectral-Entropic Uncertainty (SEU) loss by replacing it with commonly used alternatives. Replacing SEU with Dice loss or binary cross-entropy (BCE) results in noticeable performance degradation (Dice: 93.44\% and 92.03\%, respectively), confirming SEU’s advantage in capturing both spatial and uncertainty-aware features.

\textbf{Effect of Textual Guidance:}  
The role of vision-language alignment is examined by removing text prompts during inference and training. Omitting textual inputs during inference causes the Dice score to drop to 87.28\%. Eliminating textual supervision entirely by removing the MoDAB module results in a more significant performance drop (Dice: 85.15\%), reinforcing the value of language-driven guidance.

\textbf{Architectural Component Evaluation:}  
Substituting the cross-attention with point-wise addition reduces segmentation accuracy to 92.11\%, and replacing the SSMix with a linear projection yields 91.72\%. These results highlight the importance of structured attention and dynamic sequence modeling in multimodal integration.

\section{Conclusion}
In this research, we proposed a novel uncertainty-aware vision-language segmentation model designed to enhance medical image segmentation. Our model integrates visual and textual data through advanced cross-modal learning techniques, utilizing proposed key modules such as the Modality Decoding Attention Block (MoDAB) and the State Space Mixer (SSMix). These modules significantly improve segmentation accuracy by capturing both spatial and semantic information. Additionally, we proposed the Spectral-Entropic Uncertainty (SEU) Loss function, which guides the model to account for uncertainty during training, enhancing spatial precision and domain-specific visual-linguistic alignment. Comprehensive experiments on multiple medical datasets demonstrated that our model, equipped with the SEU loss, outperforms existing state-of-the-art methods in both accuracy and computational efficiency. These results underscore the potential of our approach to advance medical image analysis, offering more reliable and interpretable segmentation for clinical decision-making.

%% file: main.bib
@String(CVPR= {IEEE Conf. Comput. Vis. Pattern Recog.})

@String(ECCV= {Eur. Conf. Comput. Vis.})

@String(ICLR = {Int. Conf. Learn. Represent.})

@String(IJCAI = {IJCAI})

@String(AAAI = {AAAI})

@String(CVPR  = {CVPR})

@String(ECCV  = {ECCV})

@String(ICLR  = {ICLR})

@article{22,
  title={Cross-modal conditioned reconstruction for language-guided medical image segmentation},
  author={Huang, Xiaoshuang and Li, Hongxiang and Cao, Meng and Chen, Long and You, Chenyu and An, Dong},
  journal={IEEE Transactions on Medical Imaging},
  year={2024},
  publisher={IEEE}
}

@inproceedings{23,
  title={LGA: A Language Guide Adapter for Advancing the SAM Model’s Capabilities in Medical Image Segmentation},
  author={Hu, Jihong and Li, Yinhao and Sun, Hao and Song, Yu and Zhang, Chujie and Lin, Lanfen and Chen, Yen-Wei},
  booktitle={International Conference on Medical Image Computing and Computer-Assisted Intervention},
  pages={610--620},
  year={2024},
  organization={Springer}
}

@InProceedings{24,
        author = { Zhang, Xu and Ni, Bo and Yang, Yang and Zhang, Lefei},
        title = { { MAdapter: A Better Interaction between Image and Language for Medical Image Segmentation } },
        booktitle = {proceedings of Medical Image Computing and Computer Assisted Intervention -- MICCAI 2024},
        year = {2024},
        publisher = {Springer Nature Switzerland},
        volume = {LNCS 15009},
        month = {October},
        page = {425 -- 434}
}

@inproceedings{1,
  title={U-net: Convolutional networks for biomedical image segmentation},
  author={Ronneberger, Olaf and Fischer, Philipp and Brox, Thomas},
  booktitle={Medical image computing and computer-assisted intervention--MICCAI 2015: 18th international conference, Munich, Germany, October 5-9, 2015, proceedings, part III 18},
  pages={234--241},
  year={2015},
  organization={Springer}
}

@inproceedings{2,
  title={Unet++: A nested u-net architecture for medical image segmentation},
  author={Zhou, Zongwei and Rahman Siddiquee, Md Mahfuzur and Tajbakhsh, Nima and Liang, Jianming},
  booktitle={Deep learning in medical image analysis and multimodal learning for clinical decision support: 4th international workshop, DLMIA 2018, and 8th international workshop, ML-CDS 2018, held in conjunction with MICCAI 2018, Granada, Spain, September 20, 2018, proceedings 4},
  pages={3--11},
  year={2018},
  organization={Springer}
}

@article{3,
  title={Attention u-net: Learning where to look for the pancreas},
  author={Oktay, Ozan and Schlemper, Jo and Folgoc, Loic Le and Lee, Matthew and Heinrich, Mattias and Misawa, Kazunari and Mori, Kensaku and McDonagh, Steven and Hammerla, Nils Y and Kainz, Bernhard and others},
  journal={arXiv preprint arXiv:1804.03999},
  year={2018}
}

@article{4,
  title={nnU-Net: a self-configuring method for deep learning-based biomedical image segmentation},
  author={Isensee, Fabian and Jaeger, Paul F and Kohl, Simon AA and Petersen, Jens and Maier-Hein, Klaus H},
  journal={Nature methods},
  volume={18},
  number={2},
  pages={203--211},
  year={2021},
  publisher={Nature Publishing Group}
}

@article{5,
title = {TransUNet: Rethinking the U-Net architecture design for medical image segmentation through the lens of transformers},
journal = {Medical Image Analysis},
volume = {97},
pages = {103280},
year = {2024},
issn = {1361-8415},
doi = {https://doi.org/10.1016/j.media.2024.103280},
url = {https://www.sciencedirect.com/science/article/pii/S1361841524002056},
author = {Jieneng Chen and Jieru Mei and Xianhang Li and Yongyi Lu and Qihang Yu and Qingyue Wei and Xiangde Luo and Yutong Xie and Ehsan Adeli and Yan Wang and Matthew P. Lungren and Shaoting Zhang and Lei Xing and Le Lu and Alan Yuille and Yuyin Zhou},
}

@inproceedings{6,
  title={Swin-unet: Unet-like pure transformer for medical image segmentation},
  author={Cao, Hu and Wang, Yueyue and Chen, Joy and Jiang, Dongsheng and Zhang, Xiaopeng and Tian, Qi and Wang, Manning},
  booktitle={European conference on computer vision},
  pages={205--218},
  year={2022},
  organization={Springer}
}

@inproceedings{7,
  title={Uctransnet: rethinking the skip connections in u-net from a channel-wise perspective with transformer},
  author={Wang, Haonan and Cao, Peng and Wang, Jiaqi and Zaiane, Osmar R},
  booktitle={Proceedings of the AAAI conference on artificial intelligence},
  volume={36},
  pages={2441--2449},
  year={2022}
}

@InProceedings{8,
        author = { Liu, Jiarun and Yang, Hao and Zhou, Hong-Yu and Xi, Yan and Yu, Lequan and Li, Cheng and Liang, Yong and Shi, Guangming and Yu, Yizhou and Zhang, Shaoting and Zheng, Hairong and Wang, Shanshan},
        title = { { Swin-UMamba: Mamba-based UNet with ImageNet-based pretraining } },
        booktitle = {proceedings of Medical Image Computing and Computer Assisted Intervention -- MICCAI 2024},
        year = {2024},
        publisher = {Springer Nature Switzerland},
        volume = {LNCS 15009},
        month = {October},
        page = {615 -- 625}
}

@misc{9,
      title={U-Mamba: Enhancing Long-range Dependency for Biomedical Image Segmentation}, 
      author={Jun Ma and Feifei Li and Bo Wang},
      year={2024},
      eprint={2401.04722},
      archivePrefix={arXiv},
      primaryClass={eess.IV},
      url={https://arxiv.org/abs/2401.04722}, 
}

@inproceedings{10,
  title={Contrastive learning of medical visual representations from paired images and text},
  author={Zhang, Yuhao and Jiang, Hang and Miura, Yasuhide and Manning, Christopher D and Langlotz, Curtis P},
  booktitle={Machine learning for healthcare conference},
  pages={2--25},
  year={2022},
  organization={PMLR}
}

@inproceedings{11,
  title={TGANet: Text-guided attention for improved polyp segmentation},
  author={Tomar, Nikhil Kumar and Jha, Debesh and Bagci, Ulas and Ali, Sharib},
  booktitle={International Conference on Medical Image Computing and Computer-Assisted Intervention},
  pages={151--160},
  year={2022},
  organization={Springer}
}

@inproceedings{12,
  title={Learning transferable visual models from natural language supervision},
  author={Radford, Alec and Kim, Jong Wook and Hallacy, Chris and Ramesh, Aditya and Goh, Gabriel and Agarwal, Sandhini and Sastry, Girish and Askell, Amanda and Mishkin, Pamela and Clark, Jack and others},
  booktitle={International conference on machine learning},
  pages={8748--8763},
  year={2021},
  organization={PmLR}
}

@article{13,
  title={Large-scale domain-specific pretraining for biomedical vision-language processing},
  author={Zhang, Sheng and Xu, Yanbo and Usuyama, Naoto and Bagga, Jaspreet and Tinn, Robert and Preston, Sam and Rao, Rajesh and Wei, Mu and Valluri, Naveen and Wong, Cliff and others},
  journal={arXiv preprint arXiv:2303.00915},
  volume={2},
  number={3},
  pages={6},
  year={2023}
}

@inproceedings{14,
  title={Gloria: A multimodal global-local representation learning framework for label-efficient medical image recognition},
  author={Huang, Shih-Cheng and Shen, Liyue and Lungren, Matthew P and Yeung, Serena},
  booktitle={Proceedings of the IEEE/CVF international conference on computer vision},
  pages={3942--3951},
  year={2021}
}

@inproceedings{15,
  title={Vilt: Vision-and-language transformer without convolution or region supervision},
  author={Kim, Wonjae and Son, Bokyung and Kim, Ildoo},
  booktitle={International conference on machine learning},
  pages={5583--5594},
  year={2021},
  organization={PMLR}
}

@inproceedings{16,
  title={Lavt: Language-aware vision transformer for referring image segmentation},
  author={Yang, Zhao and Wang, Jiaqi and Tang, Yansong and Chen, Kai and Zhao, Hengshuang and Torr, Philip HS},
  booktitle={Proceedings of the IEEE/CVF conference on computer vision and pattern recognition},
  pages={18155--18165},
  year={2022}
}

@ARTICLE{17,
  author={Li, Zihan and Li, Yunxiang and Li, Qingde and Wang, Puyang and Guo, Dazhou and Lu, Le and Jin, Dakai and Zhang, You and Hong, Qingqi},
  journal={IEEE Transactions on Medical Imaging}, 
  title={LViT: Language Meets Vision Transformer in Medical Image Segmentation}, 
  year={2024},
  volume={43},
  number={1},
  pages={96-107},
  doi={10.1109/TMI.2023.3291719}}

@inproceedings{18,
  title={Ariadne’s Thread: Using Text Prompts to Improve Segmentation of Infected Areas from Chest X-ray Images},
  author={Zhong, Yi and Xu, Mengqiu and Liang, Kongming and Chen, Kaixin and Wu, Ming},
  booktitle={International Conference on Medical Image Computing and Computer-Assisted Intervention},
  pages={724--733},
  year={2023},
  organization={Springer}
}

@inproceedings{19,
  title={SLViT: Scale-Wise Language-Guided Vision Transformer for Referring Image Segmentation.},
  author={Ouyang, Shuyi and Wang, Hongyi and Xie, Shiao and Niu, Ziwei and Tong, Ruofeng and Chen, Yen-Wei and Lin, Lanfen},
  booktitle={IJCAI},
  pages={1294--1302},
  year={2023}
}

@inproceedings{20,
  title={Beyond one-to-one: Rethinking the referring image segmentation},
  author={Hu, Yutao and Wang, Qixiong and Shao, Wenqi and Xie, Enze and Li, Zhenguo and Han, Jungong and Luo, Ping},
  booktitle={Proceedings of the IEEE/CVF International Conference on Computer Vision},
  pages={4067--4077},
  year={2023}
}

@ARTICLE{21,
  author={Wu, Jianzong and Li, Xiangtai and Li, Xia and Ding, Henghui and Tong, Yunhai and Tao, Dacheng},
  journal={IEEE Transactions on Image Processing}, 
  title={Toward Robust Referring Image Segmentation}, 
  year={2024},
  volume={33},
  number={},
  pages={1782-1794},
  doi={10.1109/TIP.2024.3371348}}

@article{27,
  title={CMIRNet: Cross-Modal Interactive Reasoning Network for Referring Image Segmentation},
  author={Xu, Mingzhu and Xiao, Tianxiang and Liu, Yutong and Tang, Haoyu and Hu, Yupeng and Nie, Liqiang},
  journal={IEEE Transactions on Circuits and Systems for Video Technology},
  year={2024},
  publisher={IEEE}
}

@article{28,
  title={Language-guided Medical Image Segmentation with Target-informed Multi-level Contrastive Alignments},
  author={Li, Mingjian and Meng, Mingyuan and Ye, Shuchang and Fulham, Michael and Bi, Lei and Kim, Jinman},
  journal={arXiv preprint arXiv:2412.13533},
  year={2024}
}

@article{29,
  title={Mulmodseg: Enhancing unpaired multi-modal medical image segmentation with modality-conditioned text embedding and alternating training},
  author={Li, Chengyin and Zhu, Hui and Sultan, Rafi Ibn and Ebadian, Hassan Bagher and Khanduri, Prashant and Indrin, Chetty and Thind, Kundan and Zhu, Dongxiao},
  journal={arXiv preprint arXiv:2411.15576},
  year={2024}
}

@inproceedings{34,
  title={SGTC: Semantic-guided triplet co-training for sparsely annotated semi-supervised medical image segmentation},
  author={Yan, Ke and Cai, Qing and Zhang, Fan and Cao, Ziyan and Liu, Zhi},
  booktitle={Proceedings of the AAAI Conference on Artificial Intelligence},
  volume={39},
  pages={9112--9120},
  year={2025}
}

@article{36,
  title={Text-guided multi-stage cross-perception network for medical image segmentation},
  author={Chen, Gaoyu},
  journal={arXiv preprint arXiv:2506.07475},
  year={2025}
}

@InProceedings{covnext, 
    author    = {Liu, Zhuang and Mao, Hanzi and Wu, Chao-Yuan and Feichtenhofer, Christoph and Darrell, Trevor and Xie, Saining},
    title     = {A ConvNet for the 2020s},
    booktitle = {Proceedings of the IEEE/CVF Conference on Computer Vision and Pattern Recognition (CVPR)},
    month     = {June},
    year      = {2022},
    pages     = {11976-11986}
}

@article{medvlsm,
  title={Exploring transfer learning in medical image segmentation using vision-language models},
  author={Poudel, Kanchan and Dhakal, Manish and Bhandari, Prasiddha and Adhikari, Rabin and Thapaliya, Safal and Khanal, Bishesh},
  journal={arXiv preprint arXiv:2308.07706},
  year={2023}
}

@InProceedings{bert, 
author="Boecking, Benedikt
and Usuyama, Naoto
and Bannur, Shruthi
and Castro, Daniel C.
and Schwaighofer, Anton
and Hyland, Stephanie
and Wetscherek, Maria
and Naumann, Tristan
and Nori, Aditya
and Alvarez-Valle, Javier
and Poon, Hoifung
and Oktay, Ozan",
editor="Avidan, Shai
and Brostow, Gabriel
and Ciss{\'e}, Moustapha
and Farinella, Giovanni Maria
and Hassner, Tal",
title="Making the Most of Text Semantics to Improve Biomedical Vision--Language Processing",
booktitle="Computer Vision -- ECCV 2022",
year="2022",
publisher="Springer Nature Switzerland",
address="Cham",
pages="1--21",
isbn="978-3-031-20059-5"
}

@inproceedings{kvasir_dataset, title={Kvasir-seg: A segmented polyp dataset}, author={Jha, Debesh and Smedsrud, Pia H and Riegler, Michael A and Halvorsen, P{\aa}l and de Lange, Thomas and Johansen, Dag and Johansen, H{\aa}vard D}, booktitle={International Conference on Multimedia Modeling}, pages={451--462}, year={2020}, organization={Springer} }

@article{morozov2020mosmeddata,
  title={Mosmeddata: Chest ct scans with covid-19 related findings dataset},
  author={Morozov, S. P. and others},
  journal={arXiv preprint arXiv:2005.06465},
  year={2020}
}

@article{hofmanninger2020automatic,
  title={Automatic lung segmentation in routine imaging is primarily a data diversity problem, not a methodology problem},
  author={Hofmanninger, J. and Prayer, F. and Pan, J. and R{\"o}hrich, S. and Prosch, H. and Langs, G.},
  journal={European Radiology Experimental},
  volume={4},
  number={1},
  pages={1--13},
  year={2020},
  publisher={SpringerOpen}
}

@article{48degerli2022osegnet,
  title={OSegNet: Operational Segmentation Network for COVID-19 Detection using Chest X-ray Images},
  author={Degerli, A. and Kiranyaz, S. and Chowdhury, M. E.H. and Gabbouj, M.},
  journal={arXiv preprint arXiv:2202.10185},
  year={2022}
}

@article{loshchilov2019decoupled,
  author       = {Ilya Loshchilov and
                  Frank Hutter},
  title        = {Fixing Weight Decay Regularization in Adam},
  journal      = {CoRR},
  volume       = {abs/1711.05101},
  year         = {2017},
  url          = {http://arxiv.org/abs/1711.05101},
  eprinttype    = {arXiv},
  eprint       = {1711.05101},
  bibsource    = {dblp computer science bibliography, https://dblp.org}
}

@article{loshchilov2016sgdr,
  author       = {Ilya Loshchilov and
                  Frank Hutter},
  title        = {{SGDR:} Stochastic Gradient Descent with Restarts},
  journal      = {CoRR},
  volume       = {abs/1608.03983},
  year         = {2016},
  url          = {http://arxiv.org/abs/1608.03983},
  eprinttype    = {arXiv},
  eprint       = {1608.03983},
  bibsource    = {dblp computer science bibliography, https://dblp.org}
}

@article{mamba,
  title={Mamba: Linear-Time Sequence Modeling with Selective State Spaces},
  author={Gu, Albert and Dao, Tri},
  journal={arXiv preprint arXiv:2312.00752},
  year={2023}
}

@article{zeevi2025frequency,
  title={Enhancing Uncertainty Estimation in Semantic Segmentation via Monte-Carlo Frequency Dropout},
  author={Zeevi, Tal and Staib, Lawrence H and Onofrey, John A},
  journal={arXiv preprint arXiv:2501.11258},
  year={2025}
}

@article{antico2022evaluating,
  title={Evaluating Uncertainty Quantification in Medical Image Segmentation: A Multi-Dataset, Multi-Algorithm Study},
  author={Antico, Marta and Bruno, Giulia and Faggiano, Elena and others},
  journal={Applied Sciences},
  volume={14},
  number={21},
  pages={10020},
  year={2022},
  publisher={MDPI}
}

@article{sedai2019uncertainty,
  title={Uncertainty Guided Semi-Supervised Segmentation of Retinal Layers in OCT Images},
  author={Sedai, Suman and Mahapatra, Dwarikanath and Garnavi, Rahil},
  journal={Medical Image Analysis},
  volume={57},
  pages={226--236},
  year={2019},
  publisher={Elsevier}
}

@inproceedings{roy2019bayesian,
  title={Bayesian QuickNAT: Model Uncertainty in Deep Whole-Brain Segmentation for Structure-wise Quality Control},
  author={Roy, Abhijit Guha and Navab, Nassir and Wachinger, Christian},
  booktitle={Medical Image Computing and Computer Assisted Intervention (MICCAI)},
  pages={653--661},
  year={2019},
  organization={Springer}
}

@inproceedings{gu2021efficiently,
  title = {Efficiently Modeling Long Sequences with Structured State Spaces},
  author = {Gu, Albert and Goel, Karan and Ré, Christopher},
  booktitle = {International Conference on Learning Representations (ICLR)},
  year = {2021},
  url = {https://arxiv.org/abs/2111.00396}
}

@inproceedings{poli2023hyena,
  title = {Hyena Hierarchy: Towards Larger Convolutional Language Models},
  author = {Poli, Michael and Massaroli, Stefano and Nguyen, Eric and Fu, Daniel Y. and Dao, Tri and Baccus, Stephen and Bengio, Yoshua and Ermon, Stefano and Ré, Christopher},
  booktitle = {Conference on Neural Information Processing Systems (NeurIPS)},
  year = {2023},
  url = {https://arxiv.org/abs/2302.10866}
}

@inproceedings{dao2023flashattention2,
  title = {FlashAttention‑2: Faster Attention with Better Parallelism and Work Partitioning},
  author = {Dao, Tri},
  booktitle = {International Conference on Learning Representations (ICLR)},
  year = {2023},
  url = {https://arxiv.org/abs/2307.08691}
}
